\pdfoutput=1
\documentclass[11pt]{article}

\usepackage[]{EMNLP2023}

\usepackage{times}
\usepackage{latexsym}

\usepackage[T1]{fontenc}
\usepackage[utf8]{inputenc}

\usepackage{microtype}

\usepackage{inconsolata}

\usepackage{booktabs}
\usepackage{multirow}
\usepackage{graphicx}
\usepackage{url}

\title{Controlled Generation with Prompt Insertion for Natural Language Explanations in Grammatical Error Correction}

\author{Masahiro Kaneko$^{1,2}$ \quad
        Naoaki Okazaki$^{2}$ \\
        $^1$MBZUAI \\
        $^2$Tokyo Institute of Technology \\
        {\tt Masahiro.Kaneko@mbzuai.ac.ae} \quad
        {\tt okazaki@c.titech.ac.jp}}

\begin{document}
\maketitle
\begin{abstract}

In Grammatical Error Correction (GEC), it is crucial to ensure the user's comprehension of a reason for correction.
Existing studies present tokens, examples, and hints as to the basis for correction but do not directly explain the reasons for corrections. 
Although methods that use Large Language Models (LLMs) to provide direct explanations in natural language have been proposed for various tasks, no such method exists for GEC.
Generating explanations for GEC corrections involves aligning input and output tokens, identifying correction points, and presenting corresponding explanations consistently.
However, it is not straightforward to specify a complex format to generate explanations, because explicit control of generation is difficult with prompts.
This study introduces a method called controlled generation with Prompt Insertion (PI) so that LLMs can explain the reasons for corrections in natural language.
In PI, LLMs first correct the input text, and then we automatically extract the correction points based on the rules.
The extracted correction points are sequentially inserted into the LLM's explanation output as prompts, guiding the LLMs to generate explanations for the correction points.
We also create an Explainable GEC (XGEC) dataset of correction reasons by annotating NUCLE, CoNLL2013, and CoNLL2014\footnote{\url{https://github.com/kanekomasahiro/gec-explanation}}.
Although generations from GPT-3 and ChatGPT using original prompts miss some correction points, the generation control using PI can explicitly guide to describe explanations for all correction points, contributing to improved performance in generating correction reasons.

\end{abstract}

\section{Introduction}

Grammatical Error Correction (GEC) is the task of correcting grammatical errors in a text.
In GEC, various methods have been proposed from a wide range of perspectives, including correction performance~\cite{grundkiewicz-junczys-dowmunt-2019-minimally,chollampatt-etal-2019-cross,omelianchuk-etal-2020-gector,kaneko-etal-2020-encoder,qorib-etal-2022-frustratingly}, controlling~\cite{hotate-etal-2019-controlling,yang2022controllable,loem2023exploring}, diversity~\cite{xie-etal-2018-noising,hotate-etal-2020-generating,han2021diversity}, and efficiency~\cite{malmi-etal-2019-encode,chen-etal-2020-improving-efficiency}.
It is also important in GEC for the model to provide explanations that allow users to understand the reasons behind the corrections.
Improving explainability leads to a better judgment of whether the correction reflects the intended result, learning of grammatical knowledge, and overall enhancement of GEC systems.

\begin{figure}[!t]
  \centering
  \includegraphics[width=0.5\textwidth]{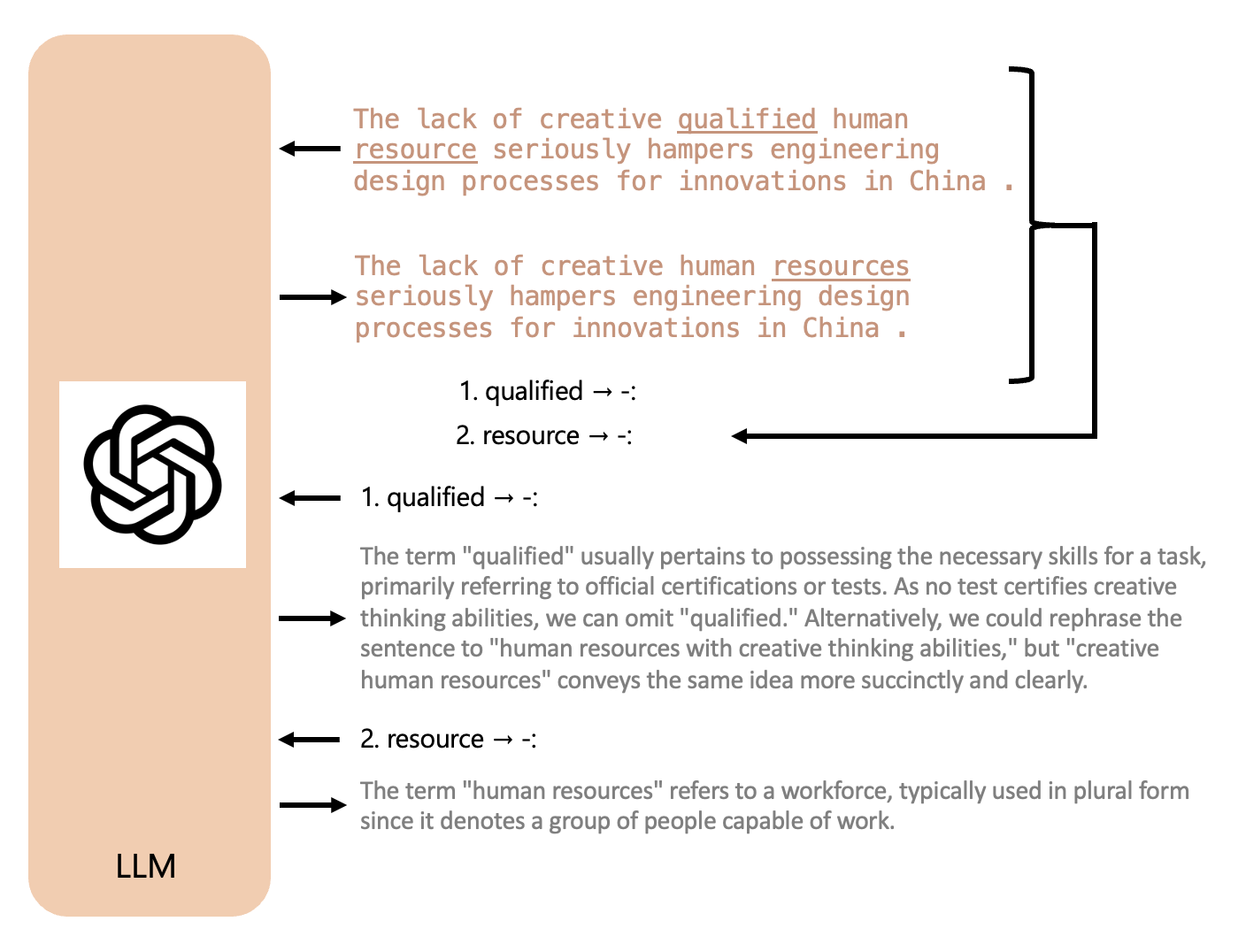}
  \caption{How to generate an explanation of the proposed method PI.}
  \label{fig:abst}
\end{figure}

\citet{kaneko-etal-2022-interpretability} introduced a method of presenting the retrieved examples as the basis for correction, in contrast to a method of retrieving data similar to the correction target from the training data set and using it for prediction.
\citet{fei2023enhancing} proposed a method that presents the token positions that are the basis of errors and error types, and showed that they are useful for learners.
\citet{nagata-2019-toward} proposed the task of generating useful hints and feedback for language learning on essays written by language learners.
This task does not necessarily generate a correction result or reason, because it is not intended for correction.
Since these existing studies do not directly explain the reason for the correction, the user must infer the reason for the correction from the system output.

Large Language Models (LLMs) such as ChatGPT~\cite{chatgpt} and GPT-3~\cite{brown2020language} have advanced language capabilities and can explain the inference reasons in natural language in various tasks~\cite{wei2022chain,wiegreffe-etal-2022-reframing,kaneko2023solving}.
With natural language, the model can directly explain the details of the inference reasons to the user.
LLMs are also effective in GEC, achieving state-of-the-art in both unsupervised~\cite{loem2023exploring} and supervised settings~\cite{kaneko2023reducing}.
Explicability in GEC first requires the alignment of input and output tokens and identifies all error and correction pairs.
Then, it is necessary to generate an explanation for each of the extracted pairs.
However, it is hard to control the generation of prompts in a specified format for GEC.
\citet{fang2023chatgpt} showed that ChatGPT improves performance by using natural language to generate step-by-step error detection and correction processes for each span.
It was found that it is difficult for ChatGPT to generate step-by-step according to the specified format with simple prompt instructions.
\citet{loem2023exploring} showed that prompting did not contribute significantly to the control of correction style for GPT-3.
%Furthermore, it was noted that it is difficult to control with simple prompt instructions in tasks other than GEC.

In this study, we introduce a method to explain the reason for correction in natural language by a controlled generation with prompt insertion (PI).
As shown in \autoref{fig:abst}, we guide LLMs to the desired format output by inserting prompts during inference.
First, LLM corrects grammatical errors in the input text.
Then, we automatically align the error and correction points from the input and output text using rules and extract error-correction pairs.
By inserting these error-correction pairs as additional prompts, we explicitly control the LLM's explanation of the reasons for all pairs.
Furthermore, we created an Explainable GEC (XGEC) dataset for explaining correction reasons in natural language by annotating NUCLE, CoNLL2013, and CoNLL2014 datasets~\cite{dahlmeier-etal-2013-building,ng-etal-2013-conll,ng-etal-2014-conll}.

In our experiments on GPT-3 and ChatGPT, we found that the original prompt-based generation resulted in pair omissions and ambiguity as to which pair the explanation was for.
On the other hand, the control of generation by PI can explicitly control the LLM to generate explanations for all the corrections, which contributes to the performance improvement of the explanation of correction reasons.

\section{PI to Generate Natural Language Explanations}

Existing methods to generate an explanation use only instructions to guide LLMs in generating explanations~\cite{wei2022chain,wiegreffe-etal-2022-reframing,chen2023models,kaneko2023solving}.
LLMs with instruction alone do not necessarily generate the explanation in the proper format covering all edits.
Our method solves this problem by inserting prompts of edits into the input during generation and explicitly guiding the LLM to generate explanations for all edits.

Specifically, we give the LLM the instruction to rewrite the input text (e.g. \textit{``What is the difference between genetic \underline{disorder} and other disorders \underline{.}''}) into grammatically correct text (e.g. \textit{``What is the difference between genetic \underline{disorders} and other disorders \underline{?}''}) and explain the corrections.
We compute the token alignment between the input and output text and extract the edits such as (\textit{``disorder $\to$ disorders''}) or (\textit{``. $\to$ ?''}).
The extracted edits are given to the LLM one by one as additional input, causing the LLM to generate an explanation corresponding to each edit.
We assign numbers to edits, such as (\textit{``1. disorder $\to$ disorders:''}) or (\textit{``2. . $\to$ ?:''}).
We assign numbers to the edits sequentially from the beginning, such as (\textit{``1. disorder $\to$ disorders:''}) or (\textit{``2. . $\to$ ?:''}).

\begin{table*}[t]
\small
\centering
\begin{tabular}{llcccccc}
\toprule
& & \multicolumn{3}{c}{XGECa} & \multicolumn{3}{c}{XGECb} \\
& & Precision & Recall & F1 & Precision & Recall & F1  \\
\midrule
\multirow{3}{*}{ChatGPT} & Post w/ PI & \textbf{83.2} & \textbf{85.5} & \textbf{84.3} & \textbf{83.9} & \textbf{84.5} & \textbf{84.2} \\
& Post w/o PI & 62.1 & 79.6 & 70.0 & 62.6 & 78.2 & 69.6 \\
& Pre w/o PI & 60.9 & 75.2 & 68.1 & 61.1 & 74.4 & 67.7 \\
\midrule
\multirow{3}{*}{GPT-3.5} & Post w/ IP & \textbf{81.2} & \textbf{83.8} & \textbf{82.4} & \textbf{82.0} & \textbf{83.0} & \textbf{82.5} \\
& Post w/o IP & 61.2 & 79.4 & 69.1 & 61.8 & 78.1 & 69.0 \\
& Pre w/o IP & 59.9 & 75.6 & 67.7 & 60.7 & 75.5 & 68.1 \\
\bottomrule
\end{tabular}
\caption{The BERTScore of GPT-3.5 and ChatGPT in generating explanations with and without PI on the XGEC test datasets.}
\label{tbl:bertscore}
\end{table*}

\section{Creating XGEC Dataset}

The XGEC dataset includes incorrect texts, correct texts, and explanations for each edit within the correct texts.
We annotated explanations for the original edits in existing GEC datasets to create training, development, and test datasets.

NUCLE~\cite{dahlmeier-etal-2013-building}, CoNLL2013~\cite{ng-etal-2013-conll}, and CoNLL2014~\cite{ng-etal-2014-conll} datasets are used as training, development, and test datasets, respectively.
NUCLE and CoNLL2013 contain only one correct text per incorrect text. 
For NUCLE and CoNLL2013, we randomly selected 362 correct texts and annotated explanations for them.
% NUCLE and CoNLL2013 have only one correct text, while CoNLL2014 has 10 correct texts for each incorrect text.
% For NUCLE and CoNLL2013, we each randomly sampled 362 correct texts and annotated the explanations.
CoNLL2014 consists of \textit{a} and \textit{b} datasets, which were created by different annotators and are commonly used to evaluating GEC models.
Consequently, we also use CoNLL2014 \textit{a} and \textit{b} for the XGEC test dataset.
To reduce the number of cases in the test dataset where our annotators disagree with the original edit, we selected edits that are widely considered appropriate by most humans.
CoNNL2014 also includes additional 8 annotations~\cite{bryant-ng-2015-far}.
Therefore, we annotated explanations for only those edits that appeared in at least 7 out of 10 correct texts within CoNLL2014 \textit{a} and \textit{b}, resulting in a total of 444 correct texts

Two native English speakers\footnote{We compensated each annotator with a payment of \$4 per explanation.} were responsible for annotating explanations for the edits.
Annotators were provided with incorrect texts, correct texts, and the corresponding edits, and they were tasked with writing an explanation for each edit in a free-writing format.
For the 10 correct texts that were not included in the annotation dataset, we provided example explanations, and the annotators used these examples as references.
In the case of NUCLE and CoNLL2013, two annotators were assigned to write explanations for each half of the correct text.
For CoNLL2014, two annotators were designated to write explanations, resulting in the creation of two references.
In total, we obtained 888 texts.

\section{Experiment}
\label{sec:experiment}

\subsection{Setting}

We used the following text as the instruction: \textit{``Correct the input text grammatically and explain the reason for each correction. If the input text is grammatically correct, only the input text should be generated as is.''}.
We used \texttt{text-davinci-003} for GPT-3.5 and \texttt{gpt-3.5-turbo-16k} for ChatGPT in OpenAI API\footnote{\url{https://platform.openai.com/docs/models/overview}}.
The number of examples for few-shot is 16.
The examples contain input texts, correct texts, edits, and explanations.
We used the ERRANT~\cite{felice-etal-2016-automatic,bryant-etal-2017-automatic}\footnote{\url{https://github.com/chrisjbryant/errant}} as the token alignment.
We automatically evaluated the performance to generate explanations with the BERTScore~\cite{zhang2019bertscore} of reference text and output text on CoNLL2014.

We compare our method with two baselines that generate explanations without inserting edit prompts.
The first baseline generates a corrected sentence and an explanation all at once.
We demonstrate the effectiveness of explicitly providing edits and generating explanations through a comparison with this baseline.
The second baseline generates an explanation all at once and then generates a corrected sentence.

The second baseline generates an explanation all at once and then generates a corrected sentence. We compare a baseline model that generates edits and explanations step by step before generating the entire corrected text, like a chain of thought, with a model that generates explanations after the entire corrected text. This demonstrates the effectiveness of generating explanations after correction.\footnote{The proposed method cannot be applied to the process of generating explanations before correction, as it requires edits extracted from correction to generate explanations.}

\begin{table}[t]
\small
\centering
\begin{tabular}{llcc}
\toprule
& & Validity & Coverage   \\
\midrule
\multirow{3}{*}{ChatGPT} & Post w/ PI & \textbf{81.5} & \textbf{100.0}  \\
& Post w/o PI  & 78.0 & 77.5 \\
& Pre w/o PI & 76.5 & 71.5 \\
\midrule
\multirow{3}{*}{GPT-3.5} & Post w/ PI & \textbf{86.5} & \textbf{100.0} \\
& Post w/o PI & 83.5 & 72.0 \\
& Pre w/o PI & 83.5 & 69.5 \\
\bottomrule
\end{tabular}
\caption{Human evaluations of GPT-3.5 and ChatGPT with and without PI on the XGEC test dataset.}
\label{tbl:human_eval}
\end{table}

\begin{table*}[t]
\small
\centering
\begin{tabular}{llcccccc}
\toprule
& & CoNLL2014 & W\&I & JFLEG \\
\midrule
\multirow{5}{*}{ChatGPT} & Pre Human & \textbf{55.2} & 51.2 & 61.7  \\
& Post Human & 54.8 & \textbf{51.5} & 61.5  \\
& Pre PI  & 54.9 & 51.7 & 61.5 \\
& Post PI & 54.7 & 49.7 & \textbf{61.8}  \\
& No explanation & 52.3 & 40.1 & 55.3 \\
\midrule
\multirow{5}{*}{GPT-3.5} & Pre Human & 54.0 & \textbf{44.2} & \textbf{57.8} \\
& Post Human & \textbf{54.5} & 44.0 & 57.3  \\
& Pre PI & 53.7 & \textbf{44.2} & 57.1 \\
& Post PI & 54.1 & 39.9 & 57.1 \\
& No explanation & 50.1 & 35.8 & 53.7 \\
\bottomrule
\end{tabular}
\caption{The GEC performance of GPT-3.5 and ChatGPT when using explanation text as examples for few-shot methods.}
\label{tbl:performance}
\end{table*}

\subsection{The Performance of Generating Explanations}

\autoref{tbl:bertscore} shows precision, recall, and F1 scores with BERTScore of GPT-3.5 and ChatGPT in generating explanations with and without PI on XGECa and XGECb datasets.
The scores of the GPT-3.5 and ChatGPT with PI are better than the models without PI in all scores on both datasets.
The performance improvement is believed to result from enhanced coverage of edits included in the explanations generated by the PI.
Moreover, it can be seen from the results of Post w/o PI and Pre w/o PI that generating explanations after correction is more effective than generating them before correction.

\section{Analysis}
\label{sec:analysis}

\subsection{Human Evaluation}

We examine the quality of LLM-generated explanations by human evaluation.
We sample 200 explanations from CoNLL2013, and four human annotators evaluate those explanations from validity and coverage perspectives.
The validity perspective refers to the accuracy and usefulness of grammatical information in LLM-generated explanations for language learners.
It is scored on three levels: 0 if the explanation for more than half of corrections is incorrect and unuseful, 1 if the explanation for more than half of corrections is correct and useful but not perfect, 2 if the explanation for all corrections is perfect.
The coverage perspective means that the LLM-generated explanation mentions all grammatical corrections.
It is scored on three levels: 0 if the explanation does not cover more than half of the corrections, 1 if the explanation covers more than half of the corrections but not all corrections, 2 if the explanation covers all corrections.

\autoref{tbl:human_eval} shows the results of validity and coverage scores from human annotators for GPT-3.5 and ChatGPT, both with and without PI.
Both the validity and coverage scores for GPT-3.5 and ChatGPT using PI are better than those not using PI.
The PI makes it clear to LLM the corrections that need to be explained, and allows for specific explanations tied to each correction, improving the quality of LLM's explanations.
The coverage scores show that by explicitly instructing correction positions using the proposed method, LLM can generate explanations that completely cover the edits.
Moreover, comparing the post-generating models and the pre-generation model demonstrates that generating an explanation before a correction has more negative effects in terms of the coverage of edits than generating an explanation after a correction.

\subsection{Impact of Explanation on GEC performance}

% Providing explanations in addition to gold answers to the LLM as few-shot examples improves performance for tasks~\cite{wei2022chain,kaneko2023solving}.
% We automatically evaluate a model's ability to generate explanations by examining the impact of explanations on GEC performance.
% It is thought that if the quality of the generated explanation is good, the GEC performance improves to the same extent as the human explanation, and if the quality is poor, the performance is not as good as the human explanation.
% We randomly sample 8 instances from the XGEC valid dataset and use them as few-shot examples.
% In order to use more generated explanatory text for evaluation, we perform random sampling for each instance of the test data to select few-shot examples.
% Add both human-written explanations and explanations generated by the PI to the few-shot examples.
% Insert explanations before and after the corrected text and compare them.

Providing explanations in addition to gold answers to the LLM as few-shot examples improves performance for tasks~\cite{wei2022chain,kaneko2023solving}.
We evaluate a model's ability to generate explanations by assessing their impact on GEC performance.
It is believed that if the quality of the generated explanation is high, the GEC performance will improve to the same extent as with human explanations. 
Conversely, if the quality is poor, the performance will not be as good as with human explanations.
We randomly sample 8 instances from the XGEC valid dataset to use as few-shot examples.
To include more generated explanatory text for evaluation, we perform random sampling for each instance in the test data to select few-shot examples. 
These examples consist of human-written explanations and explanations generated by the PI, inserted both before and after the corrected text, allowing us to compare their effectiveness, respectively.

% \autoref{tbl:performance} shows the GEC performance of GPT-3.5 and ChatGPT using explanation texts as examples of few-shot in CoNLL2014, W\&I, and JFLEG test datasets.
% Comparing no explanation results and other results, it can be seen that using explanations for examples of the few-shot method improves GEC performance.
% When comparing the results of human-authored explanatory text and text generated by PI, they both achieve nearly equivalent GEC performance.
% Therefore, The explanatory text generated by PI has the same quality as the explanatory text humans write.
% Furthermore, it can be seen that adding explanatory text before or after correction for a few-shot has little influence.

\autoref{tbl:performance} displays the GEC performance of GPT-3.5 and ChatGPT using explanatory texts as examples for few-shot learning in the CoNLL2014, W\&I, and JFLEG test datasets.
Comparing the results without explanations to the results with explanations, it is evident that using explanations as examples for few-shot learning improves GEC performance.
When comparing the results of human-authored explanatory text and text generated by the PI, both achieve nearly equivalent GEC performance. This suggests that the explanatory text generated by the PI is of the same quality as the explanatory text authored by humans.
Furthermore, it can be observed that adding explanatory text before or after correction for few-shot learning has little influence.

\section{Conclusion}
\label{sec:conclusion}

In this study, we introduce a method for generating comprehensive and high-quality explanatory text in LLMs by explicitly instructing the edits. 
Additionally, we have created the XGEC dataset for explanatory text generation.
The experimental results demonstrate that our approach, compared to methods that do not explicitly provide edits to LLMs for explanatory text generation, yields benefits in both human evaluation and automated evaluation.
In future work, we plan to investigate the impact of LLM-generated explanatory text on language learners.

\bibliography{custom}
\bibliographystyle{acl_natbib}

% \clearpage
% \appendix

\end{document}